# A Self-Organizing Neural Scheme for Door Detection in Different Environments

F.Mahmood
College of E&ME
(NUST) Peshawar Road
Rawalpindi

F.Kuwar
College of E&ME
(NUST) Peshawar Road Rawalpindi

## ABSTRACT

Doors are important landmarks for indoor mobile robot navigation and also assist blind people to independently access unfamiliar buildings.  Most existing algorithms of door detection are limited to work for familiar environments because of restricted assumptions about color, texture and shape. In this paper we propose a novel approach which employs feature based classification and uses the Kohonen Self-Organizing Map (SOM) for the purpose of door detection. Generic and stable features are used for the training of SOM that increase the performance significantly: concavity, bottom-edge intensity profile and door edges.  To validate the robustness and generalizability of our method, we collected a large dataset of real world door images from a variety of environments and different lighting conditions. The algorithm achieves more than 95% detection which demonstrates that our door detection method is generic and robust with variations of color, texture, occlusions, lighting condition, scales, and viewpoints.

## 1. INTRODUCTION

The challenge of a computer vision system is to recognize the natural structures that are designed with specific constraints on the shape, size and relative location such as doors walls and corridors. Camera based indoor navigation and way finding can assist the mobile robot to navigate in unfamiliar environment. With the ever increasing processing power of mobile hand held devices, automated obstacle avoidance can be realized on such devices to help the handicapped. Doors are one of the most widespread obstacles in indoor environment since they provide the entrance and exit points of rooms. The detection of this obstacle is necessary for space recognition, path generation and indoor map development. As a result robust real-time automatic door detection would benefit many applications, including courier robots, tour guides, patrol robots and way finding of visually impaired people.

The problem of door detection has been extensively studied in past. Much of the previous work depends on 3D information including the visual data and the distance data which includes sonar, laser and stereo-vision sensor [1-5]. These methods suffer the disadvantages of high-cost, high-power, and complexity in their systems. Murillo et al [11] apply both appearance features and shape features (color histogram) to detect doors. In [6], two neural network based door classifiers are trained by using color and shape features respectively. The main shortcoming of this approach was that it was just designed to detect the doors in familiar environment (doors of author's office building) only and would prove fail if the color of door and wall is changed. Moreover this approach used the supervised learning paradigm.  Training of such networks is a key concern because they require both the input and the desired output for each training instance and these input/target data pairs cannot easily be obtained in all situations. In [7,8] AdaBoost algorithm is employed to combine the multiple features of door such as color, door knob, door gap, doorframe, texture on bottom of a door, door width, and door concavity. By extensive experimentation we observed that some features like color and texture are not always constant in different environments. In [13] image based door detection is performed using corner and edges. Unfortunately no neural network technique is capable of door detection with sufficient accuracy in unfamiliar environments.

This paper presents a generic solution of door segmentation in single image covering different environment conditions. To address this issue, it is proposed a Kohonen Self Organizing Map (SOM) based architecture. SOM [12] is an unsupervised learning neural architecture that is a digital analogue of the brain's self-organizing capability. Research has revealed that topologically adjacent areas in the cerebral cortex perform similar cognitive functions and in these functions the relationship of physically similar stimulus is closely maintained. Likewise the nodes in Kohonen Network that represent input vectors close to each other in the input space have their weight vectors so oriented that they lie close by in the output space. In this way the weight vectors of the network will cluster the space such that they approximate the probability density function of the input vectors [12]. This capability of the SOM is employed in this paper for the purpose of door detection. The presented technique can work in all the environments including varying lighting conditions, low-contrast edges, bright reflections, textured and un-textured floors, walls and doors with similar colors. It is also suitable for real-time applications in a mobile robot.

This paper is structured as follows. The overview of the system is mentioned in section 2. Section 3 describes the Training Methodology. Section 4 presents the Experimental References are shown at the end in section 5.

## 2. SYSTEM OVERVIEW

The presented system consists of 5 steps. Firstly edges are extracted from the input grey level image using Canny Edge Detector [9], Line detection algorithm is then applied to group the edges into line segments. The next step involves the extraction of generic and significant features where some assumptions are supposed to be met. The computed features are presented in form of input vector for the training purpose of SOM. Finally the trained SOM classifier is used to perform the segmentation of the doors presented in the image. Figure 1 presents the steps used in our algorithm in chronological order. Each of the above mentioned steps are discussed in detail in the subsequent sections.

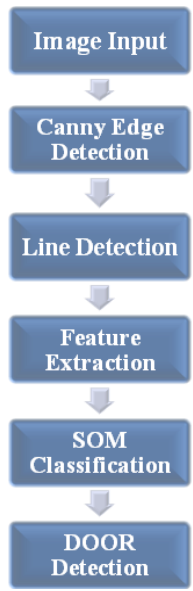

**Figure 1. Our proposed algorithm in steps**

## 2.1 Canny Edge Detection

The Canny edge detector [9] is an edge detection operator that uses a multi-stage algorithm to detect a wide range of edges in images. Its parameters allow it to be changed to recognition of edges of differing characteristics depending on the particular requirements of a given implementation. The input grey level image do not vary much in their lightness continuity i.e. the noise in them is minimal so a static Canny parameter can be readily found. Figure 2(a) shows the sample input image and Figure 2 (b) shows the output edges calculated by a Canny Edge Detector after tuning its parameters.

## 2.2 Line Detection

To group the intensity edges into straight line segments many existing approaches apply Hough transform or probabilistic Hough transform but in our experiment we found that both of them to be quite sensitive to the window size. Instead we employed the LINE DETECTION ALGORITHM developed by the Kovesi [10]. Intensity edges found by canny edge detector is given as an input as shown in fig 2(b). In the first step labeling of the edges are done. In it each unlabeled edge pixel is tracked to find the rest of connected edges and label them accordingly. This step proceeds on until the junction point is encountered. The labeled edges of the sample image are shown in figure 2(c). In the line segmenting process each labeled edge is connected by creating a virtual line from the end point of previous edge to the start point of next edge. Then the deviation i.e. the perpendicular distance of the virtual line from each point of the labeled edge is calculated. The virtual line is divides in half at that point where the calculated deviation is greater than the pre-determined threshold and this process of division continues until the deviations of the entire line segment is less than the threshold. Finally each line segment is merged with another line segment if the angle deviation and the maximum distance between their end points are within the limit. Figure 2(d) shows a final result of this algorithm in which the irrelevant and spurious edges are eliminated.

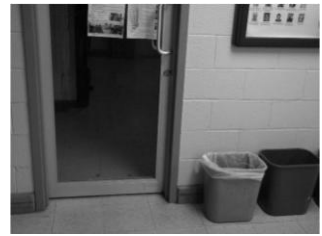 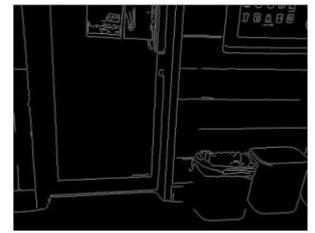

**Figure 2(a) Input image**    **Figure 2(b) output edges of canny edge detection**

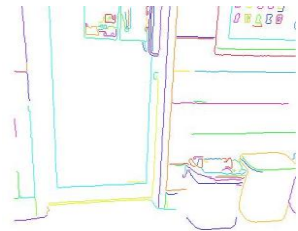 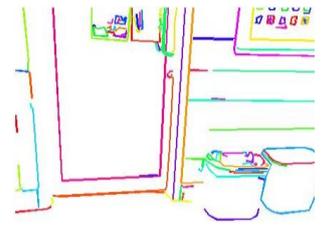

**Figure 2(c) Labeled Edges**    **Figure 2(d) Fitted line segments**

**Figure 2: Output images of different steps of Line detection algorithm**

## 2.3 Feature Extraction

The information content of a typical indoor image is huge. There is usually a need to compress information and use only that information which is needed to perform our pattern recognition task. The purpose of feature extraction is to extract all the necessary information and to ignore the redundant and useless information. In this work we used three features described in the following subsection. Our approach makes the following assumptions:

- Both door posts are visible in the image and should be nearly vertical.
- The lower portion of the door is visible in the image

The cameras must be pointed downward because of the importance of being able to see the ground to avoid obstacles. Beyond these assumptions, the presented scheme is quite forgiving: The color and pattern of the door may be same or different from the wall, Upper part of the door may be visible in the image and the door may be slightly opened or closed in the image. The purpose of utilizing the multiple features in the SOM is that the absence of one feature will not cause the door to be missed as the net-work will provide the closest possible solution because of the sufficient evidence in the other features. The selected features are described as below.

### 2.2.1 Distance between the pair of vertical line

First the horizontal distance between the two door posts is calculated. For this the pair of vertical lines is found in the image having length greater than the pre de-fined threshold and who's top extends above the vanishing point. Reflection of the doors on the ground causes un-expected long lines may lead to failure of some algorithms. In this work these unnecessary long lines were eliminated by cutting lines on the ground and optimizing the canny edge detection algorithm. In figure 3 the horizontal distance is calculated between the vertical green line pair. Let (x,y) and (x,y') be the point on right and left vertical line respectively, then the horizontal distance between these two points as calculated below:

$$Dist = |(x, y) - (x, y')| \quad (1)$$

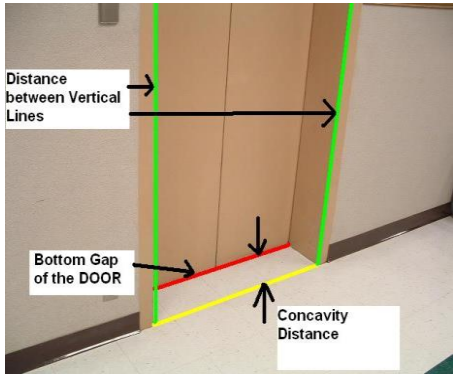

**Figure 3: Door image for explaining the used features**

### 2.2.2 Concavity

In almost all the environments doors are receded into the wall creating a concave shape for a door way. In the image the bottom edge of the door is slightly recessed from the wall/edge floor as illustrated in figure 3. The red line indicates the bottom edge of the door and the yellow line is extended portion of the wall/edge floor in front of the door edge. Let (x,y) be the point on the red line and $(x,y_d)$ be the point on the yellow line on the image, then the concavity of the door is the vertical distance between these two points as calculated below:

$$Dist = |(x, y) - (x, y_d)| \quad (2)$$

This range of the distance is 2-10 pixels depending upon the position of the robot with respect to wall i.e. the distance between them and the angle of facing of robot to the door.

### 2.2.3 Bottom-Door gap

In every environment the door is constructed with a gap below them to avoid any un-necessary friction between the door and the floor. This gap becomes darker or brighter than the surrounding area, depending upon the illumination of the room behind the door. In either case this piece of evidence is just a few pixels height but provides a vital cue for door detection. In this experiment the intensity of each pixel of the bottom door and the wall/floor edge is calculated and presented to SOM. In Figure 4 the above image shows the change of intensity values of the door gap of dark room whereas the image below demonstrates the intensity profile of the enlightened room as compared to the surrounding wall/edge floor. Position of the vertical profile is shown along x-axis whereas the pixel value is given on y-axis. This feature becomes dominant if the height of camera is 30-100 cm above the floor which proves useful in case of mobile robots.

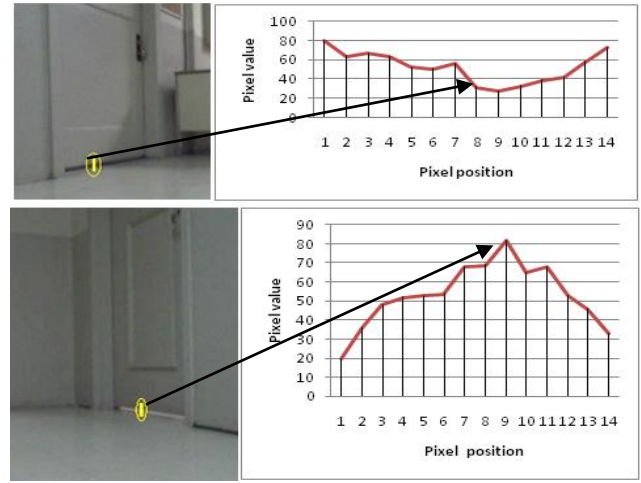

**Figure 4: Bottom edge intensity profile of the two sample images**

### 2.2.3 SOM classification

Classification through SOM plays a pivotal role in this scheme because its sole purpose is decision making i.e. differentiating the door area with the non-door area. After the SOM is trained then the feature vector formed in the process of feature extraction is given input to the SOM classifier. The output of SOM is each vertical line segment in the image i.e. the line segment extending from the top of the door to the wall/floor edge. Figure 8 shows some sample input images and the separation of door area by SOM. The SOM segments the doors presented in the image on the basis of winning neuron. The winner neuron is selected on the basis of minimum Euclidean distance from the presented features. The feature vector is first normalized to ensure that no feature dominates the classification process. The same process is done for the whole image in which each line is either given a value of zero or one to form a binary image. The diagram in Fig.5 clearly shows above mentioned process.

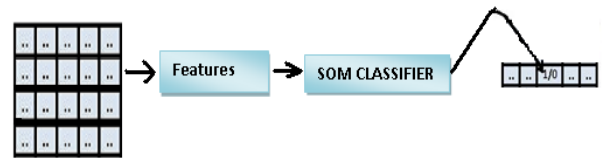

**Figure 5: Steps of SOM classifier in detecting door**

## 3 TRAINING METHOLODOGY

The proposed scheme uses SOM to detect the door area in the given input image. Firstly the SOM network has to be trained to adjust its weights, which are distributed uniformly in the beginning, to the given dataset images. For this purpose a dataset of 600 images was collected from twenty different buildings and in different illuminating conditions exhibiting a wide variety of visual characteristics. After the dataset accumulation was completed, training of the network was carried out in the sequential mode using the 'Winner Takes Most' (WTM) approach. The algorithm employed in the paper is described sequentially in Fig.6.

### 2.4.1 Initializing the weights

Prior to the training the weight vectors of all the neurons are initialized uniformly. Typically these are set to small standardized random values. The weights were initialized with random values between 0 and 1.

### 2.4.2 Calculating the Best Matching Unit

The input vectors are chosen randomly from the set of training data and presented to the lattice sequentially. Every node is examined to calculate which one's weights are mostly like the given input vector. To determine the best matching unit, one method is to iterate through all nodes and calculate the Euclidean distance between each nodes weight vector and the current input vector. The node with a weight vector closest to the input vector is tagged as BMU. The Euclidean distance is given in Eq 2 as:

$$\text{Dist} \quad \sqrt{\sum [x_i - w(i,j)]^2} \quad (3)$$

In the above equation $x_i$ is the input vector and $w(i,j)$ is the node's weight vector.

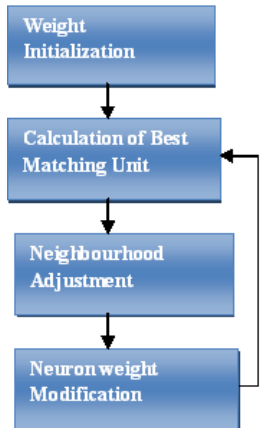

**Figure 2: Steps of training methodology of SOM**

### 2.4.3 Neuron Weight Modification

The weight of every node in the best matching unit (BMU) neighborhood including the neighborhood is modified according to the following equation

$$w_{n+1}(i,j) = w_n(i,j) + \eta(n)e(j) \quad (4)$$

Where n represents the discrete step, η is a small variable called the learning rate which decreases with time. According to the above equation the modified weight for the node is equal to the old weight ($w_n(i,j)$) plus the fraction of the difference between the input vector and the old weight. The decay of learning rate is calculated every iteration as shown in equation below

$$\eta(n) = \eta_0 e^{-\left(\frac{n}{\tau 2}\right)} \quad (5)$$

In the above equation η0 is learning rate constant and the learning rate η (n) is chosen to be a Gaussian function decreasing with the number of iterations. The used Gaussian function is in terms of covariance. The equation below depicts the difference between the input vector and the old weight of the node.

$$e(j) = x(j) - w_n(i,j) \quad (6)$$

$$w_{n+1}(k,j) = w_n(k,j) + \eta(n)h(d,\sigma)e(j) \quad (7)$$

In the above equations i and j are indices numbers, the neighborhood h(d, σ ) is a function of spread σ around the winner and distance d between the winner and the neighboring neuron k, e(j) is the difference between the winner neuron weight and the given input.

### 2.4.4 Neighborhood Adjustment

A unique feature of SOM is that the area of the neighborhood shrinks over time. In it the learning rate, neighborhood function and the spread values are updated.

$$h(d,\sigma) = h_0 e^{-\left(d^2/\sigma^2\right)} \quad (8)$$

Here h0 is the neighborhood constant and σ is given by

$$\sigma = \sigma_0 e^{(-\nu/\tau 2)} \quad (9)$$

Where σ0 is the spread constant and τ2 is the spread time constant. The initial spread constant was 4 with the spread time constant of 0.21.

### 2.4.5 Iterations

The above mentioned steps are repeated until the maximum number of iterations is reached. The training is done in two phases. In the ordering phase the spread is kept almost as large as the size of layer for quick ordering. In the convergence phase the spread and the neighborhood constants are kept small. The learning rate is kept constant for the convergence phase.

## 4 EXPERIMENTAL RESULTS

The proposed scheme was tested on video of our university hallway. It detected the entire doors successfully and showed computational time efficiency. The video feed was processed at 20 frames per second which is almost optimal for online use. The algorithm was implemented in Visual C++ and OPEN CV Library on a 1.8GHZ DELL INSPIRON 6000 Laptop with 1GB internal memory. The images were taken by an inexpensive web camera mounted 35 cm above a manually developed platform attached above P3AT mobile robot. Dataset consisting of all possible indoor environments for example all the illumination conditions from morning time till night. The dataset consisted of 600 images in which half of them were used for the training purpose. The dataset of chosen images were divided into three main categories for closely investigating the performance of SOM classifier which are: Day light, Night light and Shadows environment. Images from Day light cover the time from morning till sunset. Both the Incandescent and Inflorescent lights are covered in the night light environment. Each category contained 200 images in which half of the images were used for training purposes. All the images were 1200x1600 pixels.

First the features of door are extracted from the given input image and then presented to SOM for training purpose. After extensive experimentation we found that the network parameters are insensitive to the change in lighting environment. The learning process of adjusting the weights (taken from a random uniform distribution) to the dataset images, took approximately 3 minutes. This means that the network can be initially trained with a door dataset offline once and then implemented as an online learner which learns from a frame in the video feed once every second to ensure consistency in classification accuracy.

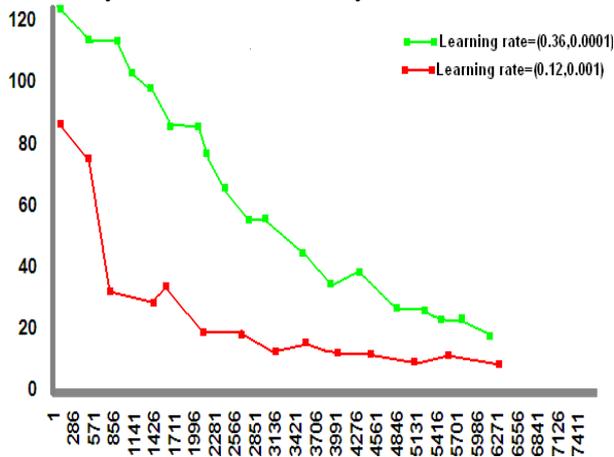

**Figure 3: Graph showing the performance of SOM with increasing iterations**

For the analysis of impact on SOM network the graph of total classification error against the number of iterations are shown in figure 7. After extensive experimentation we found that the best results showed by the SOM classifier were obtained for a learning rate of 0.12 for ordering phase and 0.001 learning rate for the convergence phase as shown by the red line in the graph. However the results of green line in the graph with different learning rates of ordering and convergence phase were almost close to the red line. The total numbers of iterations were 6272.

Table below shows the total images used for testing purpose, door images detected correctly and accuracy of each category. The accuracy of SOM classifier is evaluated on the amount of detected doors in the given input image as number of doors can be more than one in a single image. A given input image is marked as detected if at least one door in it is detected by the system. The accuracy of door detection is greatest in the night light environment and least in the shadow environment. The presence of shadow on the door causes the feature of bottom door-gap to be missed in some images thus reducing the precision of SOM classifier. The performed experiment clearly shows the evidence of the light invariability of the presented algorithm because its accuracy remains constant in changing environmental light conditions.

**Table 1: Table showing the accuracy of SOM classifier**

| Group | Day Environment | Night Environment | Shadow Environment |
|---|---|---|---|
| Images | 100 | 100 | 100 |
| Detected | 97 | 98 | 94 |
| Accuracy | 95.32 | 97.87 | 93.85 |

In addition to the accuracy, other aspect of the SOM such as initial training time, learning update time, and pattern classification time are also presented. Results are given in table 2.

**Table 2: Table showing attributes of SOM classifier**

| Attribute | SOM Classifier |
|---|---|
| Initial Update Time | 250 sec |
| Learning Update Time | 0.53 sec |
| Pattern Classification Time | 0.0012 sec |

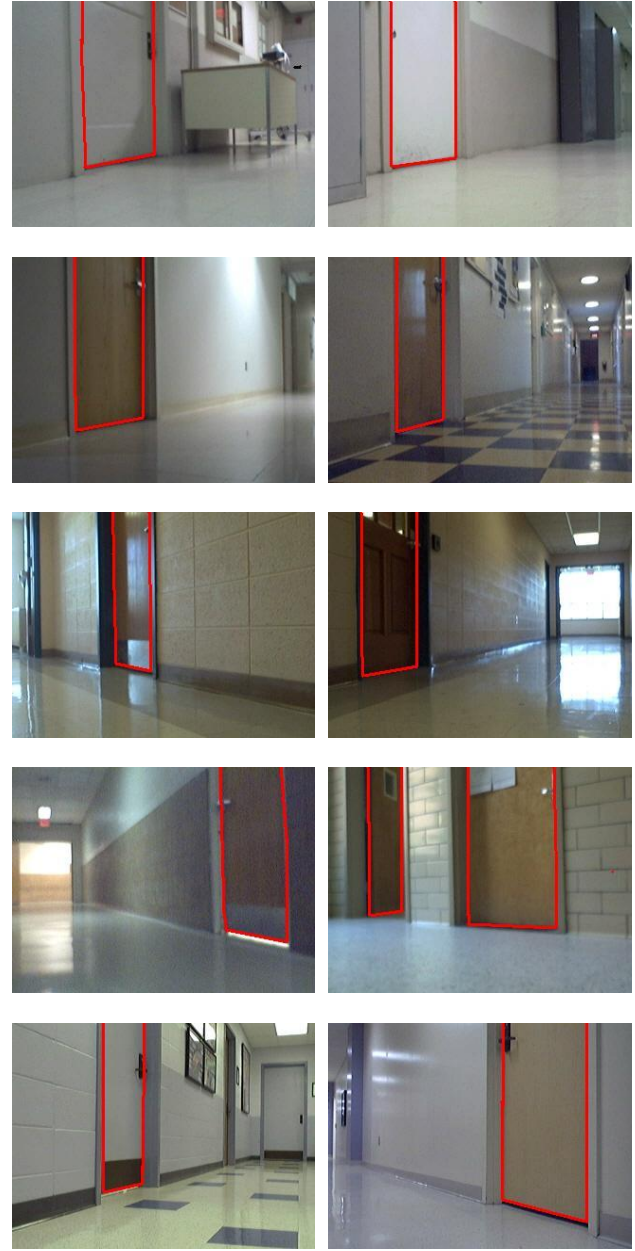

**Figure 8: Sample images of detected door**

## 5 CONCLUSION AND FUTURE WORK

In this paper we presented an approach for a camera based door detection system. By using the Self-organizing map we created a system with a detection rate more than 95% in varying indoor illumination system. Three features are used for door detection e.g. Concavity, Bottom-door gap and distance between two vertical lines. Our network took almost 0.53 sec for learning a new pattern which suggests that SOM can be used as an online learner. For the future work it would be interesting to integrate this system with an autonomous map building system. That means the robot has the capability to create a map of an unknown environment and mark the doors in it